\documentclass[lettersize, journal]{IEEEtran}
\usepackage{amsmath, amsfonts}
\usepackage{algorithm}
\usepackage{array}
\usepackage[caption = false, font = footnotesize, labelfont = rm, textfont = rm]{subfig}
\usepackage{textcomp}
\usepackage{stfloats}
\usepackage{url}
\usepackage{balance}
\usepackage{verbatim}
\usepackage{graphicx}
\usepackage{cite}
\usepackage{balance}
\usepackage{amsmath, amsfonts}
\usepackage{bm}
\usepackage{xcolor}
\usepackage{multirow}
\usepackage{makecell}
\usepackage{amsmath}
\usepackage{algpseudocode}

\begin{document}

\title{DiffRM: A Multi-Modal Generative Foundation Model for Multi-band Radio Map Construction}

\author{Zihan Huang, Tong Li, Xingzai Lv, Hua Rui, Yizhang Song, Yong Li,~\IEEEmembership{Senior~Member,~IEEE}

\thanks
{Z. Huang, T. Li, Y. Song and Y. Li are with the Beijing National Research Center for Information Science and Technology (BNRist), Department of Electronic Engineering, Tsinghua University, Beijing 100084, China. E-mail: tongli@mail.tsinghua.edu.cn.\protect 

X. Lv and H. Rui are with the State Key Laboratory of Mobile Network and Mobile Multimedia Technology, ZTE Corporation, Shenzhen 518055, China.\protect
}}


\maketitle

\begin{abstract}
The rapid proliferation of wireless applications requires accurate path loss estimation for efficient radio resource management. Radio maps provide a location-aware solution by representing received signal strength distribution over a geographical area, enabling intelligent and adaptive network management.
However, constructing high-resolution radio maps is hindered by costly sensor deployments and time-consuming driving-test data collection. 
To overcome these challenges, we propose a foundation model DiffRM for multi-band radio map construction. DiffRM leverages diffusion models to capture the probabilistic characteristics exhibited by received signal strength. Additionally, DiffRM adopts a unified structure that incorporates masking strategies to jointly perform prediction and imputation, enabling the model to generalize across diverse radio map construction tasks. Furthermore, we apply modality-specific encoders in DiffRM to embed multimodal contextual information-including environmental layouts, numerical metadata, and semantic descriptions-into latent space. This design enhances the model's adaptability across diverse scenarios, including varying terrains, sampling heights, and frequency bands.
Extensive experiments on a comprehensive radio map dataset validate its superior performance and generalization, paving the way for enhanced wireless network design and optimization.
\end{abstract}


\section{Introduction} 
\label{sec1}

The rapid development of information technologies has spawned numerous applications~\cite{li2023carbon, li2023artificial}, such as the Internet of Things (IoT), autonomous driving, and vehicular networks. To support the deployment of these innovative services, effective radio resource management is required to optimize spectral efficiency and maintain quality of service~\cite{wang2018survey, jiang20213gpp, liu2024coverage, su2025gennet}. Accurate wireless channel measurement is fundamental to this process, as it directly affects signal quality and overall network performance. As wireless networks continue to evolve, radio map (RM) technology offers an efficient solution by enabling location information-driven signal quality or path loss estimation~\cite{jiang2026pihg, zhou20263dckm, huang2026universal, jiang2025unirm}. This advancement facilitates spectrum resource allocation and enhances the adaptability of wireless networks to diverse environments. Radio maps are expected to become a key enabler in realizing the full capabilities of next-generation networks, e.g., vehicular networks and unmanned aerial vehicle (UAV) networks, supporting seamless connectivity and enabling more efficient and adaptive network management.

A radio map is a graphical representation of spatial signal power spectrum density (PSD) over the geographical environment~\cite{romero2022radio, xiao2017enhancing}. The radio map provides a visual insight into the received radio-frequency (RF) signal strength within a specific area, serving as a pivotal tool for analyzing propagation characteristics, planning network capacity, and optimizing deployment efficiency~\cite{li2015optimal, gong2023scalable}.  
In vehicular networks, where reliable and low-latency communication is crucial for safety-critical applications, radio maps are particularly essential~\cite{cao2018design}. By providing fine-grained spatial knowledge of signal propagation, radio maps assist in predicting connectivity disruptions, analyzing interference~\cite{shlezinger2020deepsic}, optimizing handoff strategies, and enhancing vehicle-to-infrastructure (V2I) and vehicle-to-vehicle (V2V) communications~\cite{roger2023deep, graff2023deep}. 

As wireless networks continue to densify with the proliferation of nodes and antennas as well as the growing demand for heterogeneous services across multiple frequency bands~\cite{shen2023toward}, constructing multi-band radio maps becomes increasingly essential. These maps enable the simultaneous analysis of signal characteristics across various frequency bands, thereby enhancing spectrum efficiency and supporting intelligent spectrum management. 
Meanwhile, the emergence of space-air-ground integrated networks has introduced new spatial dimensions, necessitating the development of three-dimensional (3D) radio maps. In contrast to conventional 2D representations, 3D radio maps provide a volumetric perspective of signal propagation, capturing variations along both horizontal and vertical axes. This capability is significant for applications such as UAV communications~\cite{li2019closed} and the deployment of aerial base stations. Therefore, developing 3D and multi-band radio maps has become a key research problem.

The conventional way to obtain radio maps is by measurements, mainly via crowd-sensing or drive-testing~\cite{hapsari2012minimization}. 
Sensing-based approach requires deploying dedicated sensors across the target area, which entails substantial costs associated with sensor deployment, calibration, and maintenance, making it economically impractical for large-scale scenarios.
Alternatively, driving tests involve collecting measurements by traversing the environment using instrumented vehicles, which reduces the need for fixed infrastructure; however, this approach is still labor-intensive and time-consuming~\cite{waheed2022deepchannel}. 
These limitations hinder the efficient and scalable construction of accurate radio maps, thereby impeding progress in both research and applications in wireless networks.


\textcolor{black}{
With the advancement of wireless propagation modeling and data-driven techniques, various approaches have been developed for radio map construction. Existing methods can be broadly categorized into two types: model-based and model-free approaches.
Model-based methods rely on predefined propagation models to estimate spatial signal distributions, such as the parallel factor analysis (PARAFAC)-based framework~\cite{fu2015factor} and the Fixed Rank Kringing (FRK)~\cite{braham2016fixed}.  
In contrast, model-free methods can be further divided into interpolation-based and learning-based approaches. Interpolation-based methods, such as inverse distance weighted (IDW) interpolation~\cite{kuo2010discriminant} and Radial Basis Function (RBF) interpolation~\cite{krumm2003minimizing}, estimate signal distributions from spatially correlated measurements. 
Learning-based methods leverage neural networks, including Convolutional Neural Networks (CNNs)~\cite{imai2019radio} and Generative Adversarial Networks (GANs)~\cite{zhang2023rme}, to capture complex propagation patterns. 
Notable works include RadioUNet~\cite{levie2021radiounet} which leverages the U-Net architecture, and a GAN-based image-driven spatial interpolation method~\cite{li2019sparsely}.}
Although these deep learning-based approaches have demonstrated better performance than traditional measurements, they still face two primary limitations~\cite{li2025generative}. 
Most existing models rely solely on unimodal inputs and fail to leverage rich multimodal information—such as environmental semantics or numerical transmitter parameters —that could provide valuable context about physical environments, transmission characteristics, and radio propagation behavior, thereby limiting the quality of radio map construction.
Additionally, these models are often tailored to a specific frequency band and sampling height, resulting in limited generalization across diverse deployment settings. Sampling height, in this context, refers to the elevation above ground level at which signal measurements are taken. When applied to scenarios involving different frequency-height combinations, they typically require substantial retraining, hindering their scalability and broader applicability. 

In this paper, we propose a multimodal-context-aware foundation model for radio map construction across multi-band scenarios. However, achieving this goal entails addressing three main challenges. 
\textbf{Firstly}, received signal strength exhibits highly stochastic behavior due to the influence of complex environmental factors~\cite{zeng2021toward}. These complex variations make it challenging to accurately capture and model the underlying signal propagation patterns, thereby posing difficulties in constructing reliable radio maps. Thus, generating high-quality radio maps requires a model that can effectively capture the stochastic nature of wireless signal propagation, preserving the underlying distribution while simultaneously capturing the complexity and variability of the surroundings. 
\textbf{Secondly}, radio map construction primarily consists of prediction and missing-value imputation, which are treated as distinct tasks with separate objectives and modeling strategies. Prediction aims to estimate radio maps in unmeasured scenarios based on building layouts and transmitter locations, while imputation focuses on reconstructing missing entries in sparsely observed radio maps without transmitter information. This separation limits the development of generalizable models, thereby highlighting the need for a foundation model that can seamlessly handle both tasks within a unified framework. \textbf{Thirdly}, radio propagation is affected by numerous environmental factors. These factors can be presented in various modalities, including building imagery, textual descriptions of weather conditions, and numerical metadata such as sampling height and frequency band. These diverse forms of contextual information provide unique and complementary insights into how radio signals behave in different scenarios. As a result, to support the construction of accurate radio maps across diverse deployment scenarios, it is essential to jointly consider and effectively integrate multimodal contextual information.

To overcome the aforementioned obstacles, we introduce DiffRM, a diffusion-based foundation model for radio map construction. 
Specifically, we employ the diffusion models as the backbone of our framework, aiming to model the probabilistic distribution of received signal strength, which solves the \textbf{first} challenge. By leveraging denoising operations, the model decomposes the original PSD distribution into a multi-step Markov chain, which allows the model to iteratively refine and construct realistic radio maps while preserving the probabilistic features present in radio propagation~\cite{luo2024rm, ho2020denoising}.
In addition, we design a unified structure that reformulates both prediction and imputation as a joint task through masking strategies. By selectively masking a subset of context during training, the model is able to learn a general inference mechanism, effectively bridging the gap between prediction and imputation tasks, which addresses the \textbf{second} challenge.
Furthermore, we apply modality-specific encoders that project heterogeneous contextual information into the latent space, which tackles the \textbf{third} challenge. Each encoder is tailored to capture the distinct semantic characteristics of its input modality, enabling the model to effectively capture the influence of multimodal context on radio propagation, allowing it to demonstrate strong adaptability across diverse scenarios with varying environmental conditions.

Our contributions can be summarized as follows:
\begin{itemize}
\item We propose a foundation model, DiffRM, for radio map construction across diverse scenarios, encompassing multiple frequency bands, various terrain types, and different sampling heights. This unified framework can capture the complex dependencies and spatial variations inherent in radio propagation based on the corresponding environmental context, achieving a strong generalization ability.

\item The proposed DiffRM employs diffusion models, which can model the probabilistic distribution of received signal strength. Also, DiffRM adopts masking strategies to unify prediction and imputation as a joint task, enabling the model to learn a general inference mechanism for diverse radio map construction tasks.
Moreover, we employ multiple encoding approaches in DiffRM, aiming to map multimodal contexts into latent representations, which enables the model to capture the propagation features revealed in these contexts, thereby enhancing the model’s adaptability across varying environments and scenarios.

\item We conduct extensive experiments on a rich radio map dataset with varying terrains, sampling heights, and frequency bands. 
\textcolor{black}{
Our proposed model, DiffRM, outperforms other baseline models by over 10\% on most metrics across both prediction and imputation tasks, demonstrating superior generalization in diverse scenarios.}
\end{itemize}

The remainder of the paper is organized as follows. Section~\ref{sec2} introduces the preliminaries. Next, we detail the proposed model in Section~\ref{sec3}. We then present numerical experiments in Section~\ref{sec4}. Section~\ref{sec5} discusses the related works. Finally, we conclude the paper in Section~\ref{sec6}.
\begin{figure*}[tb]
  \centering
  \includegraphics[width=0.9\linewidth]{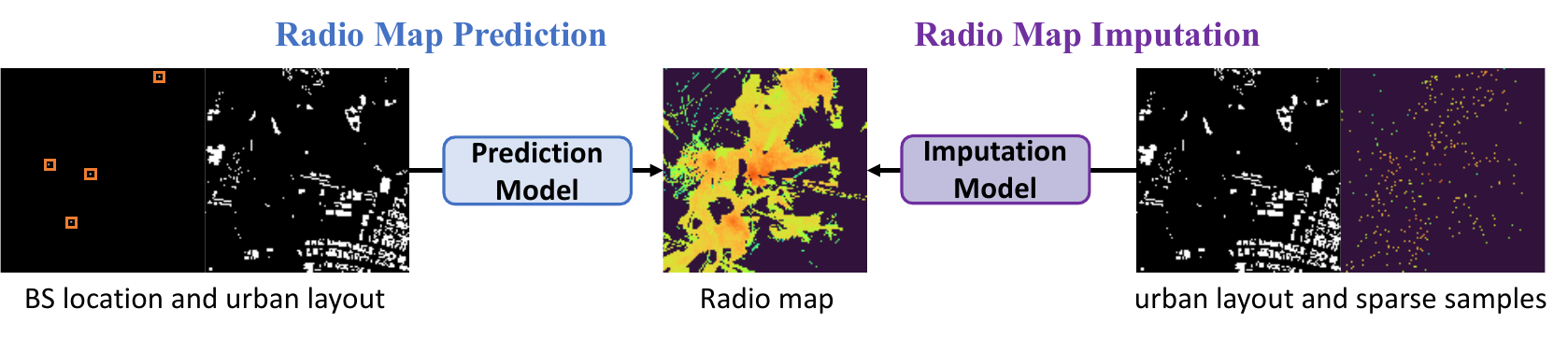}
  \caption{Difference between radio map prediction and imputation task. Prediction refers to generating signal values in entierly unmeasured regions, while imputation focuses on recovering missing values within partially observed regions.}
  \label{fig:task}
  \vspace{-2mm}
\end{figure*}

\section{Preliminaries} 
\label{sec2}

\subsection{Radio Map Construction}
Radio map construction aims to construct spatial distributions of received signal strength across a geographic area. It involves two distinct yet complementary sub-tasks as demonstrated in Figure~\ref{fig:task}, radio map prediction and radio map imputation, which differ in their input assumptions.

Radio map prediction refers to the task of generating a full-resolution radio map based on contextual information inputs, without relying on direct and actual measurements. It is applicable in restricted regions such as mountain ranges or private properties, which are inaccessible to radio map measurements because of physical constraints or security considerations~\cite {zhang2024physics}. 
Specifically, the model is provided with the environmental context of the target area, including the urban layout, transmitter location, and relevant propagation metadata. The objective is to learn a mapping from this feature space to the corresponding radio propagation field, thereby enabling the estimation of radio maps in unseen environments or configurations.

Radio map imputation, on the other hand, assumes that a sparse set of received signal strength measurements is collected within the region of interest. These sparse observed samples, along with contextual information, are supplied to the model as input features. The goal is to leverage these partial observations, together with environmental semantic priors, to reconstruct the full-resolution radio map. 
This task is effectively applied in scenarios including coverage completion from drive-testing data and signal field reconstruction for UAV or robotic path planning~\cite{zeng2021simultaneous}.

\subsection{Diffusion Models}
Diffusion models generate samples that are consistent with the original data distribution~\cite{ho2020denoising, garber2024image}. By learning to eliminate noise in a reverse process that was added sequentially in a Markovian chain during a so-called forward process, diffusion models can learn a mapping from a latent space to the original data space. 

Given a data distribution $q({\bf x}_0)$, the forward process is parameterized as:
\begin{equation}
    q({\bf x}_{1:T}|{\bf x}_0):=\prod_{t=1}^{T}q({\bf x}_{t}|{\bf x}_{t-1}),
\end{equation} 
where $q({\bf x}_{t}|{\bf x}_{t-1})=\mathcal{N}(\sqrt{1-\beta_{t}}{\bf x}_{t-1},\beta_t{\bf I})$ and $\beta_t$ is a constant adjusting the noise level.
Equivalently, ${\bf x}_t$ has a closed-form as:
\begin{equation}
    {\bf x}_t = \sqrt{\bar{\alpha}_t}{\bf x}_0+\sqrt{(1-\bar{\alpha}_t)}{\bf \epsilon}, \label{eq:DM-forward}
    \quad {\bf \epsilon}\sim\mathcal{N}({\bf 0},{\bf I}),
\end{equation}
where ${\alpha}_t:=(1-\beta_t)$, $\bar{\alpha}_t:=\prod_{s=1}^{t}{\alpha}_t$, and ${\bf \epsilon}$ is the Gaussian noise.
On the other hand, the reverse process, parameterized by $p_\theta$, learns to reconstruct $q({\bf x}_0)$ from noisy samples:
\begin{equation}
    p_{\theta}({\bf x}_{0:T}):=p({\bf x}_T)\prod_{t=1}^{T}p_{\theta}({\bf x}_{t-1}|{\bf x}_t),
    \quad
    {\bf x}_T=\mathcal{N}({\bf 0},{\bf I}).
\end{equation}
Using a particular parametrization of $p_{\theta}({\bf x}_{t-1}|{\bf x}_t)$, the reverse process can be trained using the following objective~\cite{ho2020denoising}:
\begin{equation}
    \underset{\theta}{min}\mathcal{L}(\theta):=\underset{\theta}{min}\mathbb{E}_{{\bf x}_0\sim q({\bf x}_0),{\bf \epsilon}\sim \mathcal{N}({\bf 0},{\bf I}),t}||{\bf \epsilon}-{\bf \epsilon}_\theta({\bf x}_t,t)||_2^2.
    \label{diffloss}
\end{equation}

Through this process, diffusion models effectively learn a probabilistic mapping from a Gaussian latent space to the original data distribution, allowing for the generation of high-quality samples.

\subsection{Problem Formulation}
The problem addressed in this work is to construct high-resolution radio maps for a target region by leveraging environmental context, which can be mathematically defined as follows.
Our model considers the scenario where the radio map needs to be constructed within a region of size $N_x \times N_y$ grids. Because the grid is sufficiently small, we assume that the path loss within each grid is uniform. The radio map can be represented by a matrix ${\bf y}$. Within this region, there are $N_t$ transmitters, and the environment is represented by a feature space ${\bf x}$.
Then, the dataset is defined as $\mathcal{D}=\{\mathcal{X},\mathcal{Y}\}$, where ${\bf x}\in \mathcal{X}$ and ${\bf y}\in \mathcal{Y}$.
The radio map construction task can then be denoted as $\mathcal{T}=\{\mathcal{Y}, P({\bf y}|{\bf x})\}$, where the objective is to learn a mapping from the feature space to the radio map.

For the radio map prediction task, the feature space $\mathcal{X}$ consists of (1) a building map ${\bf m}$, (2) the transmitter location $\bm{p}_{tx}$, (3) associated metadata ${K}$, and (4) a textual prompt $\tau$. The objective of this task is to find a predictive function $f(\cdot)$ to accurately estimate the radio map $\mathcal{Y}$.

In terms of the radio map imputation task, $\mathcal{X}$ contains (1) the urban map ${\bf m}$ with buildings, (2) sparse signal measurements ${\bf s}$ collected from sensors, (3) associated metadata ${K}$, and (4) a textual prompt $\tau$. The goal is to construct a network $g(\cdot)$ that reconstructs the fine-resolution radio map $\mathcal{Y}$.
\section{DiffRM} 
\label{sec3}

In this section, we present the design of our proposed DiffRM. We begin with an overview of the framework, followed by detailed descriptions of its core modules.

\begin{figure*}[tb]
  \centering
  \vspace{-2mm}
  \includegraphics[width=16cm]{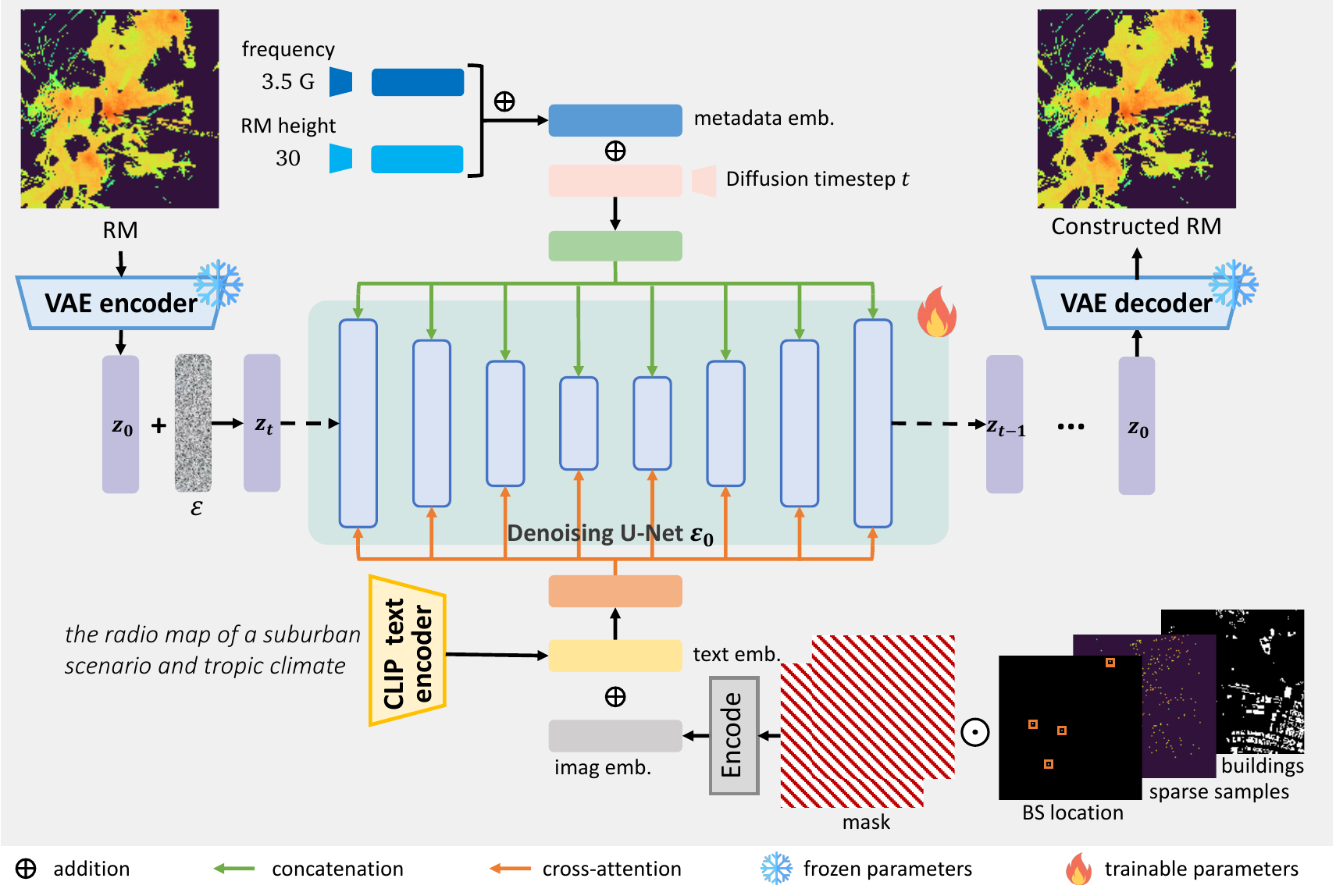}
  \vspace{-2mm}
  \caption{Overall structure of DiffRM. The model comprises three key components: (1) Latent Feature Extraction encodes radio maps into latent space; (2) Task-unified Masking integrates prediction and imputation tasks via masking strategies; and (3) Multimodal-Context-Aware Construction maps multi-modal context into latent conditions to guide diffusion-based construction. }
  \label{fig:framework}
  \vspace{-2mm}
\end{figure*}

\subsection{Framework Overview}
The overall architecture, along with each functional module of our model, is first introduced in this part. 
We leverage conditional diffusion models as the backbone, along with contextual information, to construct radio maps. 
Figure~\ref{fig:framework} illustrates the detailed structure of the proposed model, which mainly consists of three components:

\textbf{Latent Feature Extraction.} This module is responsible for encoding the raw radio map data into the latent space, enabling efficient generation via the subsequent latent diffusion process during both training and testing.

\textbf{Task-unified Masking.} This process involves unifying radio map prediction and imputation into a joint task by leveraging masking mechanisms, where a subset of context is selectively masked out during training based on the task.

\textbf{Multimodal-Context-Aware Construction.} This module is in the role of mapping the multimodal context in the feature space into the latent space, serving as conditions to guide the radio map construction via diffusion models.

\subsection{Latent Feature Extraction}
To lower the computational demands of training diffusion models and capturing high-level spatial propagation patterns, following the previous study~\cite{rombach2022high}, we utilize an autoencoder model that learns a space perceptually equivalent to the original space, preserving essential properties while facilitating the subsequent training process.

We train a variational autoencoder (VAE)~\cite{kingma2013auto} to encode the raw radio map image into the latent space. Given a radio map ${\bf y}$, the encoder $\mathcal{E}$ encodes ${\bf y}$ into a latent representation ${\bf z}_0=\mathcal{E}({\bf y})$, where noise is added to the ${\bf z}_0$ in the forward process of diffusion models according to~\eqref{eq:DM-forward} for training. After the noise is removed via the reverse process, the decoder $\mathcal{D}$ of the VAE is utilized to reconstruct the radio map from the latent denoised representation ${\bf \hat{z}}_0$. 
By using the VAE, the denoised diffusion process can simply remove the noise vector added to the latent representation ${\bf z}_0$ rather than the noise added to the original radio map image, thereby reducing the dimensionality of the diffusion models to improve training efficiency.

It is worth highlighting that the VAE training process is independent of the subsequent training of diffusion models, where the VAE only uses raw data ${\bf y}$ to train the encoder $\mathcal{E}$ and decoder $\mathcal{D}$ in an autoencoder style. Then, the parameters of the VAE remain constant throughout the subsequent training procedure of the diffusion models.

\subsection{Task-unified Masking}
To facilitate a foundation model capable of handling both radio map prediction and imputation, we introduce masking strategies that modulate the environmental context inputs according to the underlying task.

A key distinction between tasks lies in the treatment of transmitter location features $\bm{p}_{tx}$. In the imputation setting, we mask the entire location information from the input. This design enforces the model to rely solely on environmental layout and observed samples for learning the PSD distribution patterns, while excluding any positional priors. Conversely, for the prediction task, the location information is fully retained, providing essential spatial dependencies for radio map construction. We adopt a task-dependent binary mask matrix ${\bf M}_{tx}$ to control the visibility of location information:
\begin{equation}
    \bm{ p}_{tx} = {\bf M}_{tx} \odot \bm{p}_{tx}, \quad {\bf M}_{tx}=\begin{cases}

\mathbf{0}, & \text{imputation} \\

\mathbf{1}, & \text{prediction}
\end{cases}
.
\end{equation}

In addition to location-based masking, we adopt a random masking strategy over the radio map samples to simulate varying levels of observed sampling sparsity. Suppose the observation percentage is $r\in [0,1]$, representing the proportion of received signal strength measurements available within the target area. For the imputation task, we consider $r>0$ where the model is required to reconstruct the full-resolution radio map ${\bf y} $ from a sparsely observed subset of measurements ${\bf s}$. In contrast, in prediction tasks, we set $r=0$, indicating that no measurements are available in the target region, the model exclusively relies on environmental context and surrounding information to make constructions:
\begin{equation}
    {\bf M}_{\bf s}=\mathbb{I}({\bf U}[0,1]<r), \quad {\bf s} = {\bf M}_{\bf s} \odot {\bf y},
\end{equation}
where ${\bf M}_{\bf s}$ is the binary mask matrix and ${\bf U}[0,1]$ denotes uniform random variables.

\subsection{Multimodal-Context-Aware Construction}
The feature space $\mathcal{X}$ involves different types of context, including numerical metadata, textual prompts, and urban map images. In this module, we employ encoding techniques to map multimodal contexts to a latent space, which can efficiently capture the characteristics of the corresponding radio map displayed in contextual information.

\subsubsection{Numerical Metadata Encoding}
Radio maps are commonly associated with metadata, including their working frequency band, sampling height, and various other numerical information that affect radio propagation in the real world. 
Specifically, we adopt the same sinusoidal timestep embedding used in diffusion models to encode each metadata item $k_j \in {K}$ into a latent representation:
\begin{equation}
\begin{split}
    & Project(k_j,2i)=sin(k_j\Omega^{-\frac{2i}{d}}), \\
    & Project(k_j,2i+1)=cos(k_j\Omega^{-\frac{2i}{d}}),
\end{split}
\end{equation}
where $i$ is the index of feature dimension in the encoding, $d$ is the vector dimension, and $\Omega=10000$ is a constant. For every metadata information, a distinct MLP encodes the projected metadata value identically as the diffusion timestep $t$:
\begin{equation}
{\bf k}_j=MLP([Project(k_j,0),\dots,Project(k_j,d)]),
\end{equation}
\begin{equation}
{\bf t}=MLP([Project(t,0),\dots,Project(t,d)]),
\end{equation}
where ${\bf k}_j\in \mathbb{R}^D$ represents the learned MLP embedding, $\bf{t}$ is the diffusion timestep embedding, and $D$ is the embedding dimension.
After that, the diffusion timestep embedding ${\bf t}$ and all of the metadata vectors are added together as ${\bf k} = {\bf k}_1+{\bf k}_2+\dots+{\bf t}$, resulting in ${\bf k}$ as the final numerical metadata embedding.

\subsubsection{Textual prompt encoding}

In addition to numerical metadata, other information that affects radio propagation and is therefore associated with radio maps, such as terrain types and weather conditions, is difficult to depict using structured numerical features. To capture a broader environmental context, textual prompts are introduced to our model. Specifically, the prompt $\tau$ is first tokenized and then encoded to latent representations $\boldsymbol{\tau}’$ via the CLIP model~\cite{radford2021learning}. 
CLIP is a multimodal model designed to align textual and visual information within a shared embedding space. Its text encoder transforms the input prompt into a fixed-length embedding while maintaining high-level semantic meaning. The textual embedding serves as an expressive representation of the environmental context, enabling the subsequent radio map construction process to incorporate scenario-specific information that is otherwise difficult to extract. Other modalities typically capture geometric or physical features but fall short in conveying high-level contextual attributes. Textual descriptions, on the other hand, provide a flexible approach to encoding nuanced environmental characteristics, which can significantly enhance the accuracy and adaptability of radio map prediction in diverse real-world scenarios.

\subsubsection{Conditional Radio Map Construction}
To achieve conditional radio map construction based on multimodal context, we employ a conditional diffusion generative model using an attention-based UNet structure. 
For numerical metadata representation ${\bf k}$, it is concatenated with the output of intermediate layers in the U-Net architecture, allowing the model to integrate numerical information directly during the denoising process. 
In contrast, for visual information (including building map ${\bf m}$, transmitter location ${\bm p}_{tx}$ or sparse measurements ${\bf s}$) and textual representation $\boldsymbol{\tau}'$, they are first projected into the same dimension and concatenated together to obtain fused representations ${\bf C}$. Then ${\bf C}$ is injected into the intermediate layers of the denoising model via a cross-attention mechanism:
\begin{align}
     \bf{Q}&=\bf{W}_Q\cdot\phi(\bf{z}_t), \\
     \bf{K}&=\bf{W}_K\cdot \bf{C}, \\
     \bf{V}&=\bf{W}_V\cdot \bf{C}, \\
     Attention(\bf{Q},\bf{K},\bf{V})&= softmax(\frac{\bf{Q}\bf{K}^T}{\sqrt{d}})\cdot \bf{V},
\end{align}
where $\phi(\bf{z}_t)$ is the flattened operator applied on the output from the UNet layer, $\bf{W}_Q$, $\bf{W}_K$, and $\bf{W}_V$ are all trainable matrices. By applying this approach, the denoising process can be extensively guided by semantic visual and textual context.

To summarize, we first encode the radio map image ${\bf y}$ to a latent representation ${\bf z}_0=\mathcal{E}({\bf y})$ using the VAE encoder. Next, Gaussian noise is added to ${\bf z}_0$ following~\eqref{eq:DM-forward}. Then, the metadata embedding ${\bf k}$ obtained from metadata encoding techniques, CLIP-embedded textual representation $\boldsymbol{\tau}'$, as well as the visual information embedding $\bf{C}$ are served as conditions and passed through the diffusion model $\epsilon_{\theta}({\bf z}_t; {\bf k}, \boldsymbol{\tau}' {\bf C})$ to predict the added noise. Finally, the VAE decoder upsamples the denoised latent ${\bf \hat{z}}_0$ to full resolution, constructing the synthetic radio map $\hat{\bf y}=\mathcal{D}({\bf \hat{z}}_0)$.
\section{Numerical Experiments} 
\label{sec4}

\subsection{Dataset Description}

\begin{figure*}[t]
    \centering
    \subfloat[Dense urban]{\includegraphics[width = 1.5in]{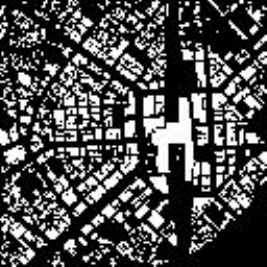}}\label{fig:a}
    \hfil
    \subfloat[Ordinary urban]{\includegraphics[width = 1.5in]{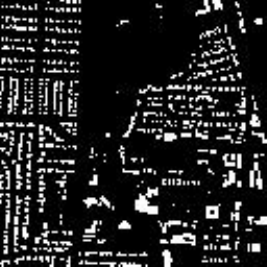}}\label{fig:b}
    \hfil
    \subfloat[Suburban]{\includegraphics[width = 1.5in]{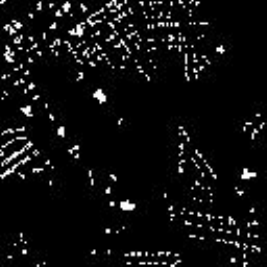}}\label{fig:c}
    \hfil
    \subfloat[Island]{\includegraphics[width = 1.5in]{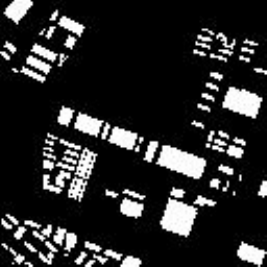}}\label{fig:d}
    \caption{SpectrumNet data examples of different scenarios. (a) dense urban scenario with high-rise buildings and complex topology; (b) ordinary urban scenario with medium-density buildings and regular street patterns; (c) suburban scenario with sparse buildings and open spaces; (d) island scenario with limited urban infrastructure.}
    \label{fig:dataset_example}
\end{figure*}

In this work, we adopt the publicly available 3D ray-tracing-based radio map dataset, SpectrumNet~\cite{zhang2024generative}, to evaluate the performance of DiffRM against baseline methods.
\textcolor{black}{The SpectrumNet dataset comprises 15,300 real-world building maps obtained from OpenStreetMap~\cite{haklay2008openstreetmap}, covering 11 diverse terrain scenarios. Each area spans 1.28 km $\times$ 1.28 km in the horizontal plane and is discretized with a spatial resolution of 10 m.
For each area, radio maps are provided at three different heights, reflecting the radio maps for ground nodes, nodes of building heights, and aerial nodes, respectively. The dataset also includes radio maps at five frequency bands ranging from 150 MHz to 22 GHz, covering most of the working frequencies of wireless networks, illustrating full-band radio propagation.
The dataset construction incorporates realistic environmental variations, including different climates, building materials, and terrain properties, which are considered during the radio map generation process, thereby improving its representativeness of real-world wireless propagation environments.
Details about the dataset are presented in Table~\ref{tab:dataset}, and Figure~\ref{fig:dataset_example} illustrates an example of the dataset. 
Before training, all the radio map values are normalized to the range of $[-1,1]$ to facilitate efficient model training. 
Additionally, each radio map is associated with two textual cues: terrain type and climate. Terrain type is chosen from 11 categories defined in the SpectrumNet dataset, and climate is chosen from 3 categories. 
In addition to SpectrumNet, we also evaluate the model on two additional datasets: RadioMapSeer~\cite{levie2021radiounet} and PMNet~\cite{lee2023pmnet}, to assess generalization under different propagation conditions. RadioMapSeer contains 700 city-scale radio maps at 5.9 GHz, while PMNet consists of over 11,000 city-scale samples at 2.5 GHz.}

\textcolor{black}{
For both tasks, all datasets are divided into training, validation, and testing sets with a ratio of 7:1:2. The partition is performed at the area level, such that all samples from the same geographic region are assigned to the same subset, thereby ensuring no overlap between different subsets and preventing spatial data leakage.}


\begin{table}[t]
\footnotesize
\begin{center}
\vspace{-2mm}
    \caption{SpectrumNet Dataset Statistics.} \label{tab:dataset}
    \begin{tabular}{c|c}
    \hline
    \textbf{Properties} & \textbf{SpectrumNet} \\
    \hline
    \hline
    \textbf{\# Urban maps} & {15,300}\\
    \hline
    \textbf{Image size} & {128 $\times$ 128} \\
    \hline
    \textbf{Spatial resolution} & {10 m} \\
    \hline
    \multirow{2}{*}{\textbf{Scenario types}} & {dense urban, ordinary urban, rural, suburban, forest,} \\
    & {mountainous, desert, grassland, island, ocean, lake}\\
    \hline
    \textbf{Sample heights} & {1.5 m, 30m, 200m} \\
    \hline
    \textbf{Frequencies} & {150 MHz, 1.5 GHz, 1.7 GHz, 3.5 GHz, 22 GHz} \\
    \hline
    \textbf{Climate types} & {tropics, subtropics, temperate zone} \\
    \hline
    \end{tabular}
    \vspace{-2mm}
\end{center}
\end{table}

\subsection{Evaluation Metrics}
To comprehensively evaluate the quality of radio map construction, we adopted five metrics commonly used in previous studies~\cite{levie2021radiounet}~\cite{wang2024radiodiff}, which are outlined below:

\textbf{Mean Square Error (MSE)}. This metric is calculated by averaging the squared discrepancies between the original and generated images' pixel intensities, shown below:
\begin{equation}
    MSE=\frac{1}{n}\sum_{i=1}^{n}({\bf y}_i-{\bf \hat{y}}_i)^2,
\end{equation}
where $n$ is the test set size, ${\bf y}$ and ${\bf \hat{y}}$ are ground truth and estimated radio map, respectively. 
The normalized MSE (NMSE) is a scaled form of MSE, which normalizes the MSE concerning the variance of the ground truth values, and root MSE (RMSE) is the square root of MSE, which are defined as below:
\begin{align}
    \rm{NMSE} & = \frac{\frac{1}{n}\sum_{i=1}^n ({\bf y}_i-{\bf \hat{y}}_i)^2}{\sum_{i=1}^n {\bf y}_i^2}, \\
    \rm{RMSE} & = \sqrt{MSE}.
\end{align}

\textbf{Structural Similarity Index Measurement (SSIM)}. This metric assesses the preservation of structural information to emphasize the precision of structural detail construction, making it aligned with human visual perception:
\begin{equation}
    SSIM(x,y)=\frac{(2\mu_x\mu_y+C_1)(\sigma_{xy}+C_2)}{(\mu_x^2+\mu_y^2+C_1)(\sigma_x^2+\sigma_y^2+C_2)},
\end{equation}
where $x$, $y$ correspond to two different input images, $\mu_x$, $\sigma_x$, $\sigma_{xy}$ denote the mean and variance of $x$ and the covariance of $x$ and $y$ respectively, $C_1$ and $C_2$ are constants to stabilize the division when the denominator becomes very small.

\textbf{Peak Signal-to-noise Ratio (PSNR)}. This metric measures the maximum possible power of a signal and the distortion introduced by interference noise that affects the fidelity of radio map construction. It is used particularly to determine the quality of edge detail in the constructed radio maps, which is computed as follows:
\begin{equation}
    PSNR=10\cdot log_{10}(\frac{r^2}{MSE}),
\end{equation}
where $r$ is the maximal variation in the input radio map image data.

\subsection{Baselines}
To evaluate the proposed model, we select the following baselines, which represent state-of-the-art techniques in the radio map construction task, as listed below.
\begin{itemize}
    \item \textbf{Autoencoder~\cite{teganya2021deep}}. This model develops a fully convolutional deep completion autoencoder architecture to exploit the structure of radio maps effectively.
    \item \textbf{conditional GAN~\cite{mirza2014conditional}}. This model is the conditioned version of the generative adversarial network, using contextual information to guide data generation.
    \item \textbf{RadioUNet~\cite{levie2021radiounet}}. It is a sampling-free NN-based method, where a convolutional U-Net is used as the backbone NN, discovering the relationship between radio maps and urban geometry.
    \item \textbf{PMNet~\cite{lee2024scalable}}. This model is a CNN-based framework, where the symmetric encoder-decoder module is adopted to facilitate efficient context propagation, and atrous convolution is employed to handle scale variations and capture broader context in urban map data.

    \item \textbf{UNetDCN~\cite{jaensch2024radio}}. Similar to RadioUNet, this model also utilizes the U-Net framework as its backbone. It applies a deformable convolution layer to replace some of the convolutions in an otherwise fairly standard UNet structure, making it easier to propagate contextual information in arbitrary directions.

    \item \textbf{PEFNet~\cite{jiang2024physics}}. This model integrates mathematical equations that govern the distribution of electromagnetic fields with machine learning. The U-Net-based neural network acts as a surrogate for the physical knowledge-driven numerical solver in radio map construction.

    \item \textbf{RadioDiff~\cite{wang2024radiodiff}}. It models radio map construction as a conditional generative problem and employs an attention U-Net with an adaptive fast Fourier transform module as the backbone diffusion model-based network to improve the dynamic environmental features extraction capability.
\end{itemize}

We conduct all model experiments on a GPU server equipped with 80 GB of memory and an NVIDIA GeForce RTX 2080 Ti GPU. 
\textcolor{black}{DiffRM is trained using the AdamW optimizer with a weight decay of $1\times10^{-4}$. The initial learning rate is set to $5\times10^{-5}$ and decayed to a minimum of $5\times10^{-6}$ using a polynomial schedule. The batch size is 32, and gradient accumulation is employed over 8 steps, resulting in an effective batch size of 256. Training is performed for 50,000 steps, and the exponential moving average (EMA) is applied starting from step 10,000 and updated every 10 steps to stabilize the training process.}
\textcolor{black}{For baseline methods, all models are trained and evaluated on the same dataset and data splits as DiffRM. To ensure a fair comparison, we adopt consistent training settings across all models, including the same batch size and number of training epochs. In addition, the key hyperparameters of each baseline are tuned within reasonable ranges, and the best performance on the validation set is reported.}

\begin{table*}[t]
\small
\renewcommand{\arraystretch}{1.26}   
\begin{center}
\caption{Radio map construction performance comparison on the SpectrumNet dataset.} 
\label{tab:overall}
\resizebox{1.0\linewidth}{!}{
\begin{tabular}{c|c|c|c|c|c|c|c|c|c|c}
\hline
{\textbf{Task}} & \multicolumn{5}{c|}{\textbf{Prediction}} & \multicolumn{5}{c}{\textbf{Imputation}} \\
\hline
{\textbf{Method}} & \textbf{MSE} & \textbf{NMSE} & \textbf{RMSE} & \textbf{SSIM} & \textbf{PSNR} & \textbf{MSE} & \textbf{NMSE} & \textbf{RMSE} & \textbf{SSIM} & \textbf{PSNR} \\
\hline
\hline
AutoEncoder & 0.0405$\pm$0.0009 & 0.1075$\pm$0.0028 & 0.2012$\pm$0.0038 & 0.2640$\pm$0.0065 & 6.2139$\pm$0.11 & 0.1387$\pm$0.0010 & 0.7312$\pm$0.0032 & 0.3688$\pm$0.0042 & 0.1732$\pm$0.0062 & 5.9213$\pm$0.12 \\
\hline
cGAN & 0.0347$\pm$0.0010 & 0.0922$\pm$0.0025 & 0.1863$\pm$0.0035 & 0.3030$\pm$0.0060 & 6.7802$\pm$0.10 & 0.1244$\pm$0.0011 & \underline{0.5837$\pm$0.0028} & 0.3056$\pm$0.0039 & 0.2029$\pm$0.0058 & 6.7802$\pm$0.10 \\
\hline
RadioUNet & 0.0298$\pm$0.0007 & 0.1714$\pm$0.0030 & 0.1863$\pm$0.0036 & 0.5106$\pm$0.0055 & 11.7690$\pm$0.09 & 0.1439$\pm$0.0005 & 0.8270$\pm$0.0035 & 0.4093$\pm$0.0045 & 0.1400$\pm$0.0065 & 4.0409$\pm$0.10 \\
\hline
PMNet & 0.0285$\pm$0.0009 & 0.1674$\pm$0.0028 & 0.1675$\pm$0.0033 & 0.5257$\pm$0.0052 & 13.7149$\pm$0.10 & 0.1453$\pm$0.0010 & 0.8268$\pm$0.0033 & 0.3802$\pm$0.0042 & 0.1404$\pm$0.0063 & 6.5437$\pm$0.11 \\
\hline
UNetDCN & 0.0278$\pm$0.0010 & 0.1606$\pm$0.0026 & 0.1667$\pm$0.0031 & \underline{0.5302$\pm$0.0050} & \underline{13.8320$\pm$0.11} & 0.1396$\pm$0.0008 & 0.7787$\pm$0.0030 & 0.3736$\pm$0.0039 & 0.1483$\pm$0.0060 & 6.6003$\pm$0.12 \\
\hline
PEFNet & 0.0294$\pm$0.0006 & \underline{0.0851$\pm$0.0024} & 0.1714$\pm$0.0034 & 0.4365$\pm$0.0058 & 9.3575$\pm$0.10 & 0.1757$\pm$0.0009 & 0.7183$\pm$0.0031 & 0.4001$\pm$0.0043 & \underline{0.3143$\pm$0.0056} & \underline{9.9005$\pm$0.09} \\
\hline
RadioDiff & \underline{0.0274$\pm$0.0003} & 0.0944$\pm$0.0023 & \underline{0.1595$\pm$0.0029} & 0.2920$\pm$0.0059 & 6.6793$\pm$0.08 & \underline{0.0366$\pm$0.0003} & 0.5964$\pm$0.0027 & \underline{0.1913$\pm$0.0033} & 0.2110$\pm$0.0061 & 6.8931$\pm$0.09 \\
\hline
\hline
DiffRM & \textbf{0.0269$\pm$0.0002} & \textbf{0.0589$\pm$0.0015} & \textbf{0.1279$\pm$0.0022} & \textbf{0.6014$\pm$0.0040} & \textbf{16.5147$\pm$0.05} & \textbf{0.0097$\pm$0.0003} & \textbf{0.0523$\pm$0.0013} & \textbf{0.0715$\pm$0.0018} & \textbf{0.6319$\pm$0.0036} & \textbf{18.7736$\pm$0.07} \\
\hline
Improv.(\%) & {1.82} & {30.79} & {19.81} & {13.43} & {19.39} & {73.49} & {91.04} & {62.62} & {101.04} & {89.62} \\
\hline
\end{tabular}
}
\vspace{-2mm}
\end{center} 
\end{table*}

\begin{table*}[t]
\small
\renewcommand{\arraystretch}{1.2}
\begin{center}
\caption{Radio map construction performance comparison on RadioMapSeer and PMNet datasets.}
\vspace{-3mm}
\label{tab:add_overall}
\resizebox{0.8\linewidth}{!}{
\begin{tabular}{c|c|ccc|ccc}
\hline
\multirow{2}{*}{\textbf{Dataset}} & \multirow{2}{*}{\textbf{Method}}
& \multicolumn{3}{c|}{\textbf{Prediction}} 
& \multicolumn{3}{c}{\textbf{Imputation}} \\
\cline{3-8}
& & \textbf{RMSE} & \textbf{SSIM} & \textbf{PSNR} 
  & \textbf{RMSE} & \textbf{SSIM} & \textbf{PSNR} \\
\hline
\hline

\multirow{3}{*}{RadioMapSeer}
& PEFNet 
& 0.1602$\pm$0.0030 & 0.4521$\pm$0.0055 & 9.9832$\pm$0.09 
& 0.3725$\pm$0.0038 & 0.3320$\pm$0.0052 & 10.4545$\pm$0.08 \\
\cline{2-8}
& RadioDiff 
& {0.1483$\pm$0.0026} & 0.3054$\pm$0.0054 & 7.1514$\pm$0.07 
& {0.1768$\pm$0.0029} & 0.2387$\pm$0.0056 & 7.4274$\pm$0.08 \\
\cline{2-8}
& DiffRM 
& \textbf{0.1185$\pm$0.0020} & \textbf{0.6287$\pm$0.0035} & \textbf{17.3603$\pm$0.05} 
& \textbf{0.0642$\pm$0.0015} & \textbf{0.6598$\pm$0.0032} & \textbf{19.8531$\pm$0.06} \\

\hline

\multirow{3}{*}{PMNet}
& PEFNet 
& 0.1558$\pm$0.0028 & 0.4683$\pm$0.0052 & 10.4290$\pm$0.08 
& 0.3604$\pm$0.0035 & 0.3451$\pm$0.0050 & 10.9832$\pm$0.08 \\
\cline{2-8}
& RadioDiff 
& {0.1427$\pm$0.0025} & 0.3186$\pm$0.0051 & 7.6352$\pm$0.07 
& {0.1689$\pm$0.0027} & 0.2513$\pm$0.0054 & 7.8848$\pm$0.08 \\
\cline{2-8}
& DiffRM 
& \textbf{0.1096$\pm$0.0019} & \textbf{0.6425$\pm$0.0032} & \textbf{17.1239$\pm$0.05} 
& \textbf{0.0587$\pm$0.0014} & \textbf{0.6712$\pm$0.0030} & \textbf{19.4338$\pm$0.06} \\
\hline
\end{tabular}
}
\vspace{-3mm}
\end{center}
\end{table*}

\begin{figure*}[t]
    \centering
    \subfloat[ground truth]{\includegraphics[width = 1.8in]{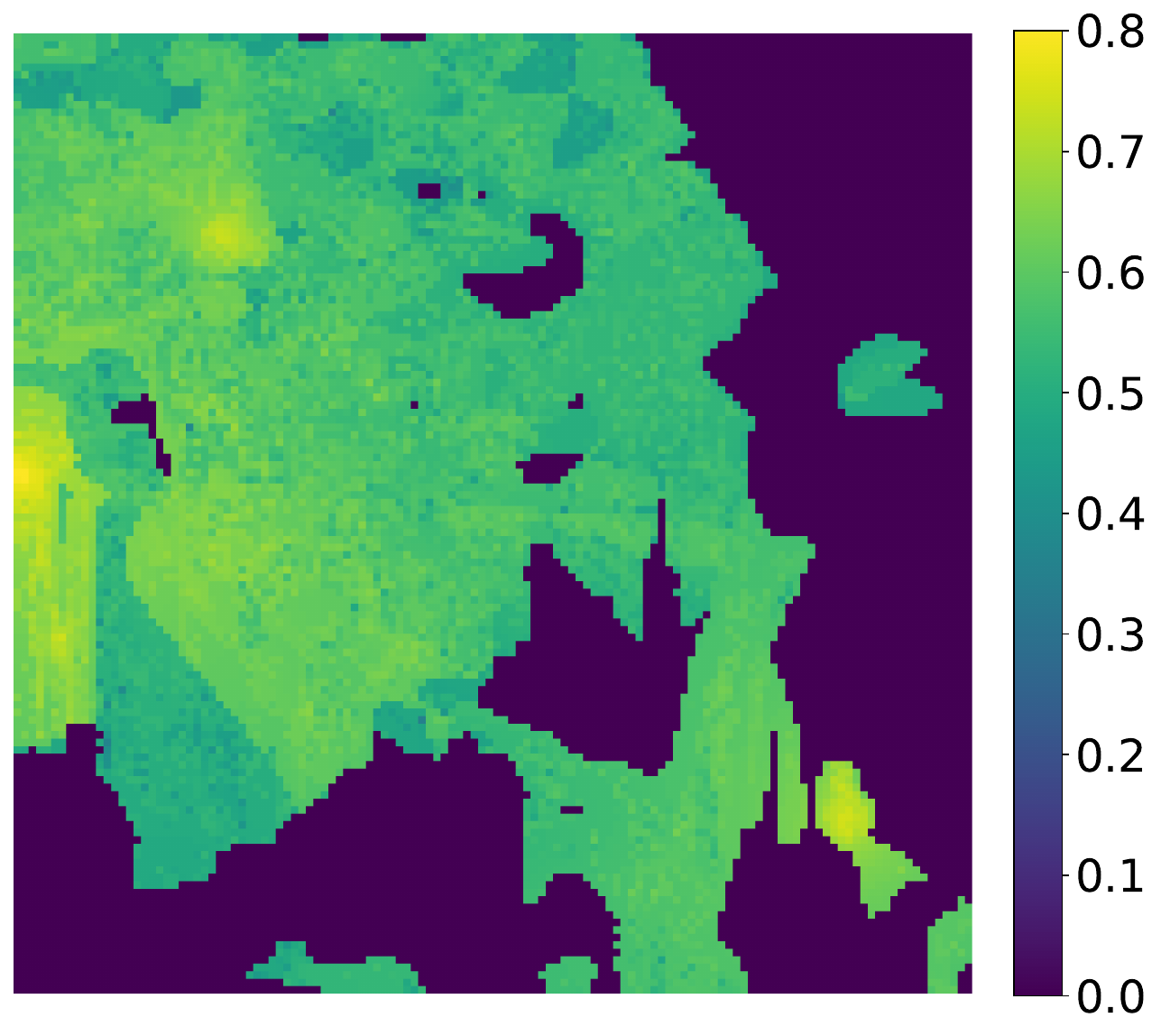}\label{fig: overall_a}}
    \hfil
    \subfloat[RadioDiff]{\includegraphics[width = 1.8in]{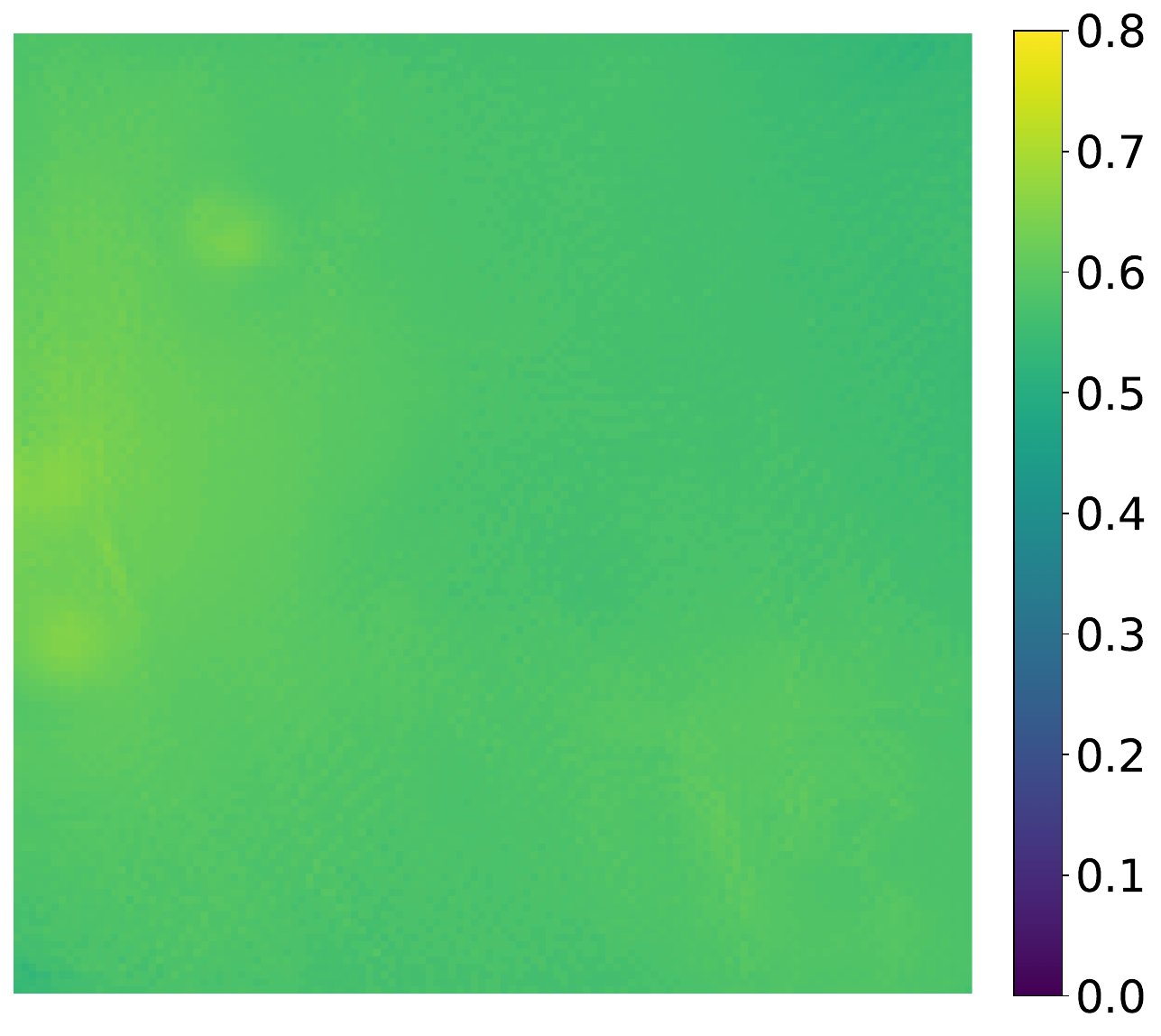}\label{fig: overall_b}}
    \hfil
    \subfloat[DiffRM]{\includegraphics[width =  1.8in]{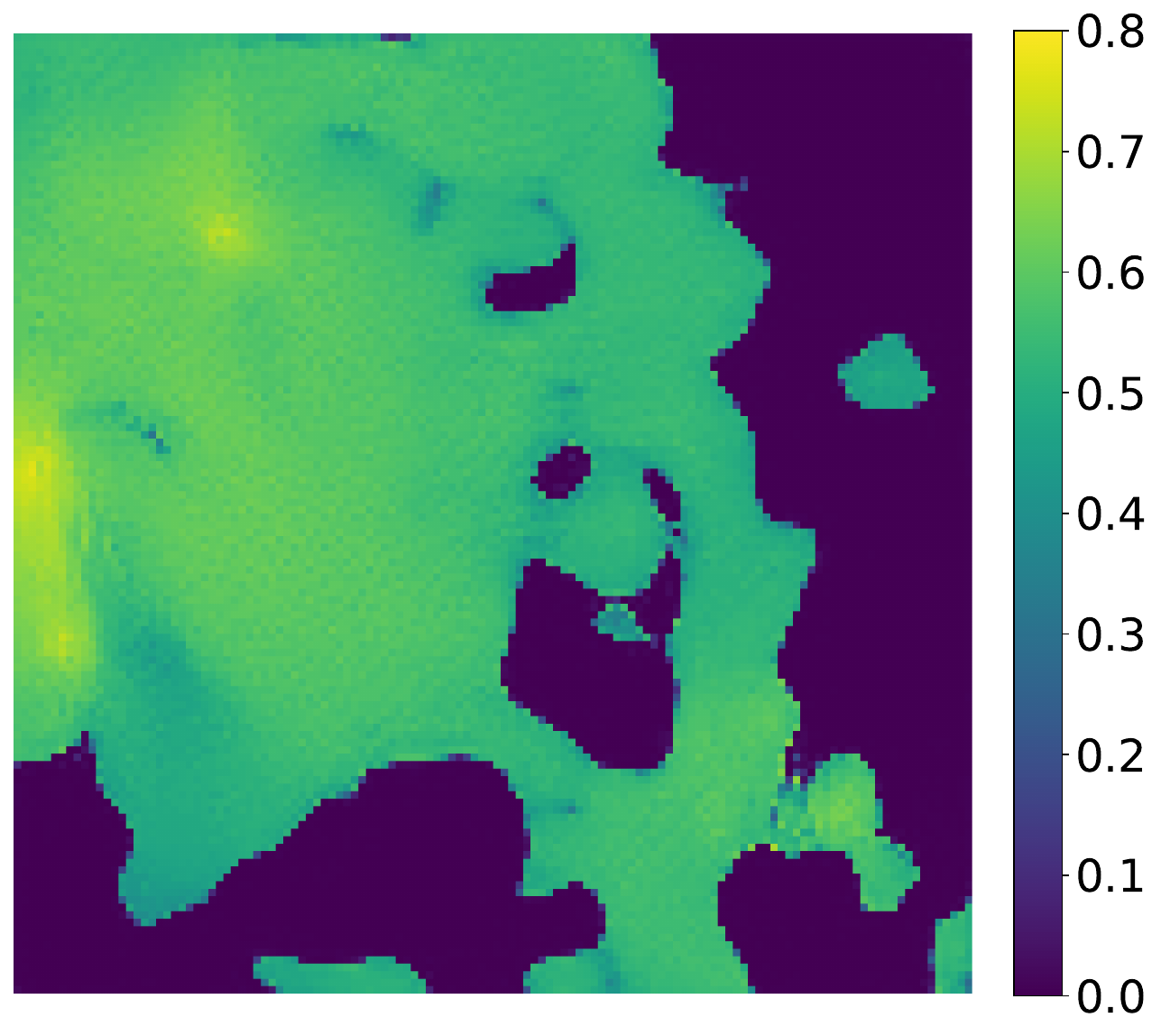}\label{fig: overall_c}}
    \hfil
    \subfloat[ground truth]{\includegraphics[width =  1.8in]{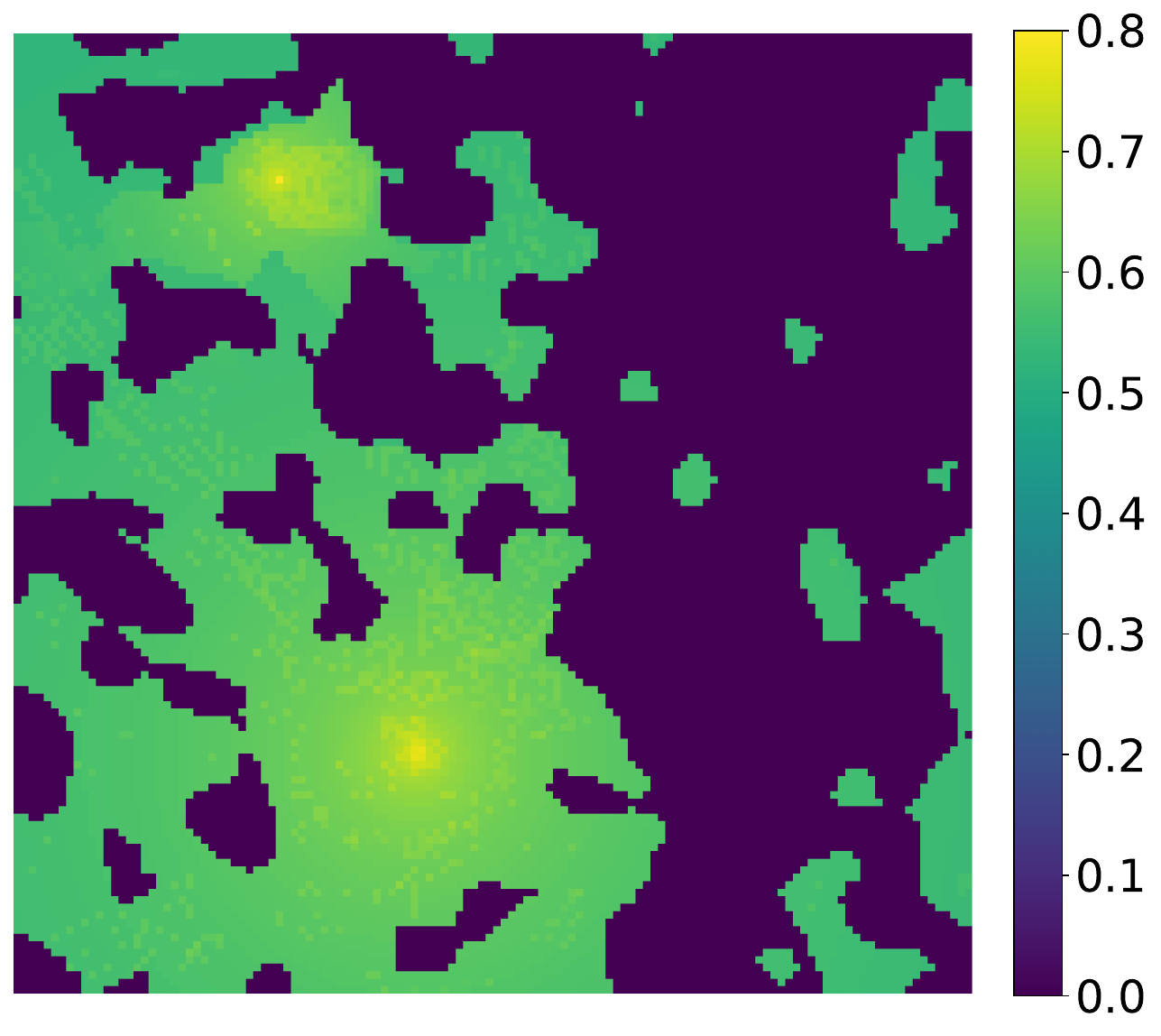}\label{fig: overall_d}}
    \hfil
    \subfloat[RadioDiff]{\includegraphics[width =  1.8in]{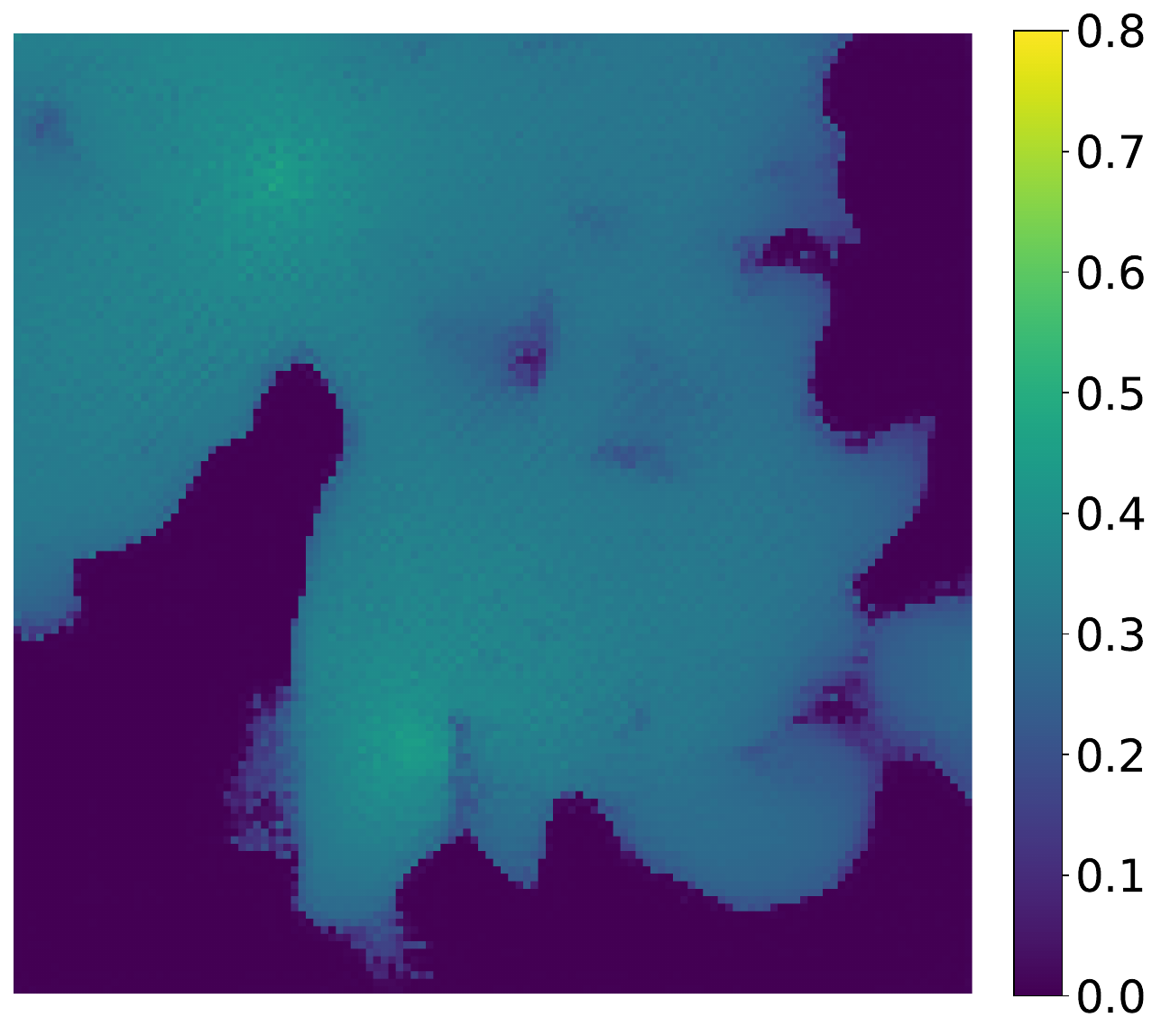}\label{fig: overall_e}}
    \hfil
    \subfloat[DiffRM]{\includegraphics[width =  1.8in]{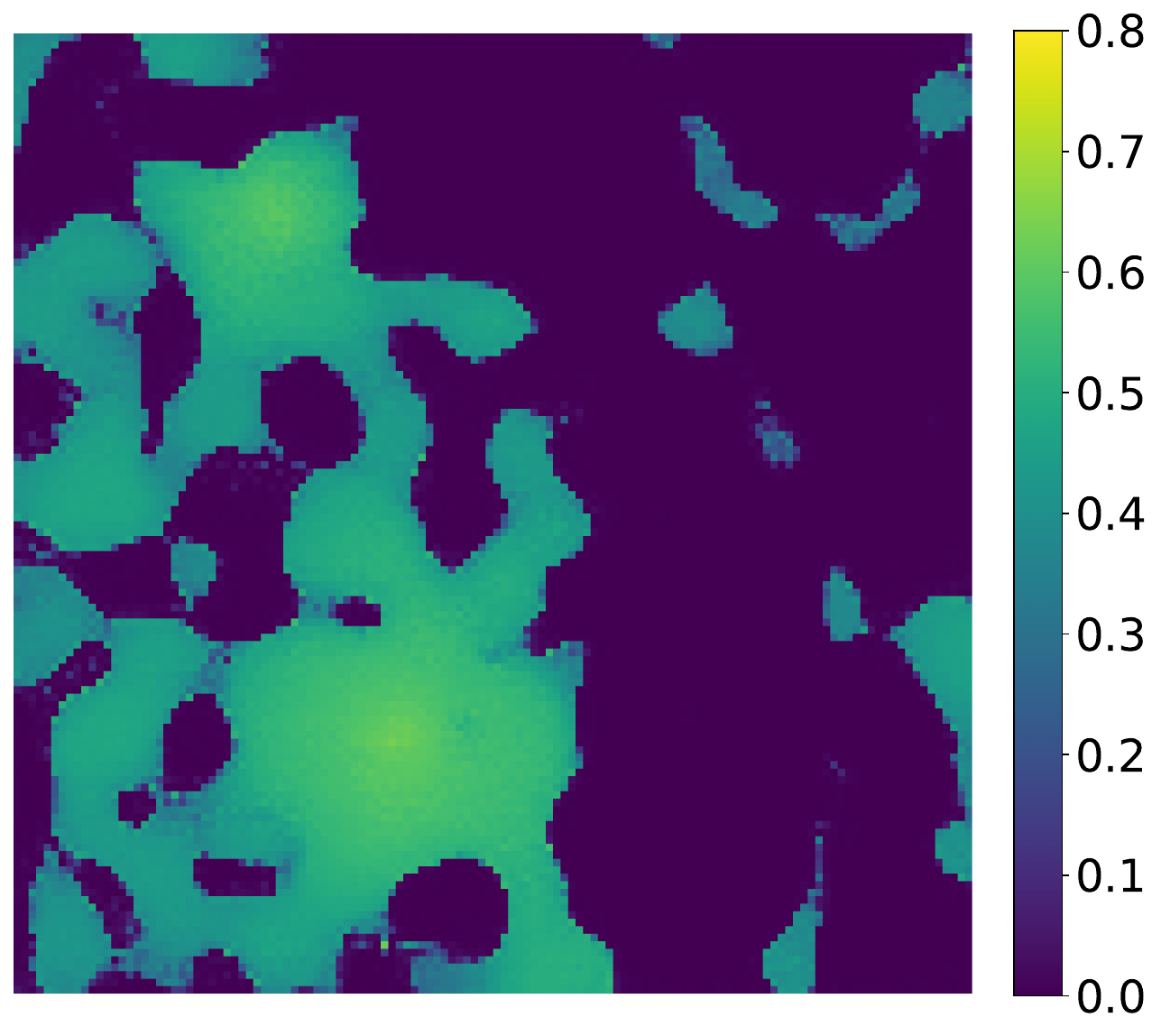}\label{fig: overall_f}}
    \caption{The comparison of constructed radio map on SpectrumNet dataset. Subfigures (a)–(c) and (d)–(f) correspond to two different scenarios. \textcolor{black}{In each group, (a) and (d) show the ground truth, (b)and (e) are the results of a representative baseline RadioDiff, (c) and (f) are from DiffRM.}}
    \label{fig: overall}
\end{figure*}

\begin{figure*}[tb]
  \centering
  \includegraphics[width=16cm]{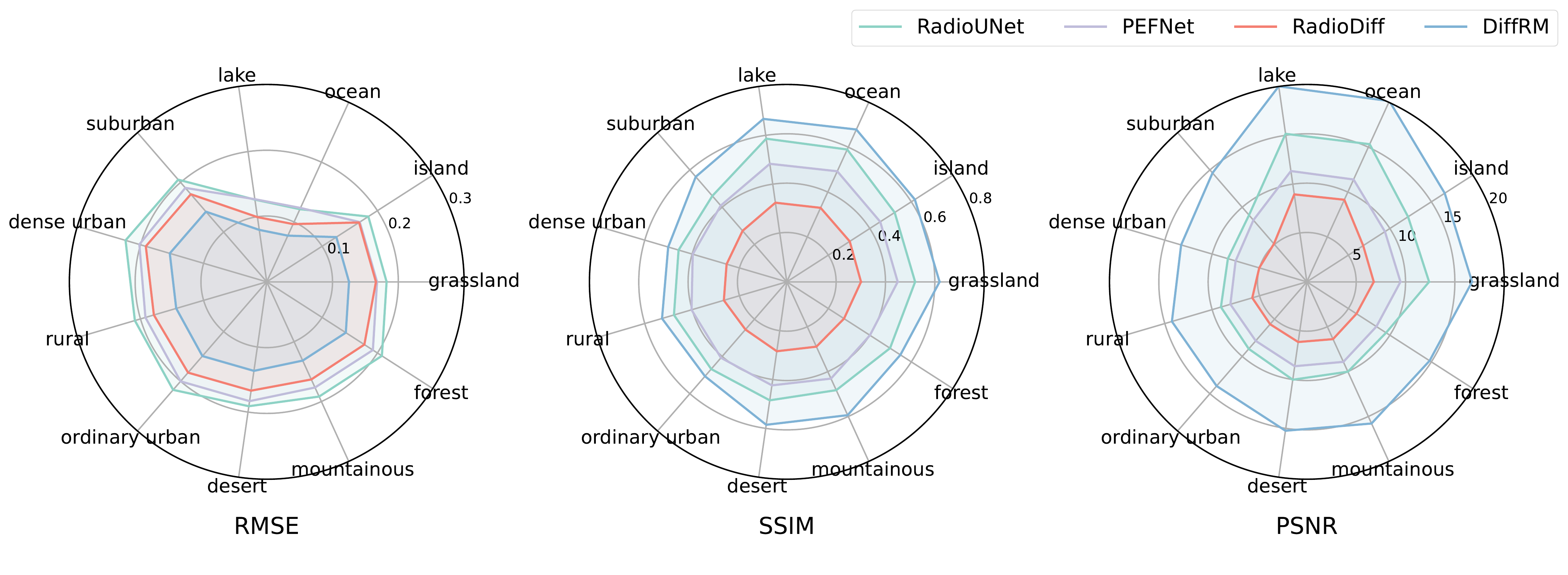}
  \caption{Radio map prediction performance in various scenarios under RMSE, SSIM, and PSNR metrics. A smaller enclosed area in RMSE indicates better performance, while larger areas in SSIM and PSNR indicate better construction quality.}
  \label{fig:radar}
\end{figure*}

\subsection{Overall Performance}
In this part, we evaluate the performance of DiffRM with other baseline methods in both radio map prediction and imputation tasks. In terms of the imputation task, we randomly select 165 sampling points as the observed sample inputs for radio map construction, resulting in a sampling rate of 1\%, without access to transmitter location information. Results are presented in Table~\ref{tab:overall}.
We observe that RadioDiff far exceeds the performance in radio map construction in both tasks compared to other baseline models, as it leverages the diffusion model architecture. As a probabilistic generative framework, the diffusion models enable RadioDiff to better capture complex probabilistic properties and generate high-quality radio maps across diverse scenarios.
However, RadioDiff still falls short of the performance achieved by DiffRM. Our model incorporates richer multimodal context, including numerical metadata, textual prompts, and urban map images, which allows the model to more effectively capture environmental information, thereby enhancing its generalization capability for radio map construction in complex scenarios across different terrains, heights, and frequency bands.

We visualize the performance of RadioDiff and DiffRM on ordinary urban and island scenarios selected from the dataset, as shown in Figure~\ref{fig: overall}. Compared to RadioDiff, our model demonstrates a clear improvement, with more accurate boundary delineation due to the shadowing effect of obstacles, as well as a closer alignment with the ground truth in reflecting the spatial variations of path loss intensity. This further indicates that DiffRM is more powerful in capturing fine-grained radio propagation patterns under different environmental conditions, which is consistent with the results displayed in Table~\ref{tab:overall}.
\textcolor{black}{
Moreover, the average inference time of DiffRM is approximately 25 ms per sample, indicating its feasibility for deployment in vehicular network scenarios.
}
\textcolor{black}{
To further evaluate the generalization capability of DiffRM, we also conduct experiments on two other datasets, RadioMapSeer and PMNet. Compared to SpectrumNet, both datasets exhibit different environmental characteristics and propagation conditions.
The results in Table~\ref{tab:add_overall} show that DiffRM maintains superior performance across both datasets, further demonstrating its robustness under diverse propagation conditions.}

\subsection{Discussions on Generalization} 
In this part, we validate the generalization of DiffRM across various terrain scenarios, sampling heights, and frequency bands.

\subsubsection{Performance in Various Terrain Scenarios}
We first assess the performance of radio map construction across different terrain scenarios, with results presented in Figure~\ref{fig:radar}. 
We can observe that the efficacy of all methods varies significantly depending on the type of terrain. When the terrain is flat and free of obstacles, such as buildings or vegetation, radio propagation is relatively predictable and straightforward, as it exhibits clearer radio patterns. Thus, the constructed radio maps in these scenarios, such as ocean and lake scenarios, tend to achieve better performance.
In contrast, scenarios with complicated terrain features, such as urban and suburban areas, present challenges for accurate radio map construction. These environments often involve irregular surfaces and numerous scatterers and reflectors, resulting in complex multipath shadowing and reflection effects, which lead to relatively lower performance in radio map construction.

\subsubsection{Performance at Different Heights}
We next evaluate radio map construction at various heights. Table~\ref{tab:height} displays the performance of different models at three distinct heights provided in the dataset. 
We can see that as the altitude increases, the performance of all four models gradually improves, and the aerial network at a height of $200$ m achieves the best performance. This disparity is attributed to the more detailed obstacle features and the more decisive influence of shadowing and reflecting effects from buildings on the terrestrial network at $1.5$ m, introducing more complexity compared to the aerial network. As a result, generating accurate radio maps for terrestrial scenarios at lower heights is more challenging.

\begin{table*}[tb]
\small 
\renewcommand{\arraystretch}{1.26}  
\begin{center}
    \caption{Radio map prediction performance at different heights.} 
    \vspace{-2mm}
    \label{tab:height}
    \begin{tabular}{c|c|c|c|c|c|c|c|c|c}
    \hline
    \textbf{Height} & \multicolumn{3}{c|}{\textbf{1.5 m}} & \multicolumn{3}{c|}{\textbf{30 m}} & \multicolumn{3}{c}{\textbf{200 m}} \\
    \hline
    \hline
     \textbf{Method} & {RMSE} & {SSIM} & {PSNR} & {RMSE} & {SSIM} & {PSNR} & {RMSE} & {SSIM} & {PSNR}\\
    \hline
    \textbf{RadioUNet} & {0.2434} & {0.4682} & {8.3810} & {0.1805} & {0.5101} & {11.3203} & {0.1682} & {0.5313} & {12.4628} \\
     \hline
    \textbf{PEFNet} & {0.2314} & {0.4103} & {6.8921} & {0.1703} & {0.4287} & {9.2199} & {0.1422} & {0.4535} & {11.0032} \\
     \hline
    \textbf{RadioDiff} & {0.2298} & {0.2479} & {5.3892} & {0.1503} & {0.2942} & {6.4021} & {0.1376} & {0.3148} & {7.9321} \\
    \hline
    \textbf{DiffRM} & {0.2256} & {0.5429} & {15.9677} & {0.1090} & {0.5610} & {16.1056} & {0.0543} & {0.6612} & {16.9063} \\
    \hline
    \end{tabular}
\end{center} 
\end{table*}

\subsubsection{Performance in Multiband Scenarios}
Furthermore, we verify the performance of radio map construction across different frequencies. Table~\ref{tab:frequency} shows the experimental results for five distinct frequency bands.
\textcolor{black}{
It can be observed that radio map construction at 22 GHz generally exhibits greater challenges compared to intermediate frequency bands, particularly 3.5 GHz. This difference may be attributed to the propagation characteristics of high-frequency signals, which are more susceptible to environmental obstructions and exhibit limited diffraction and penetration capabilities. Consequently, radio propagation patterns at higher frequencies can be more sensitive to complex environmental conditions, posing additional challenges for accurate radio map construction.}

\subsubsection{\textcolor{black}{Performance under different observation densities}}
\begin{figure*}[t]
    \centering
    \subfloat[RMSE]{\includegraphics[width = 2.0in]{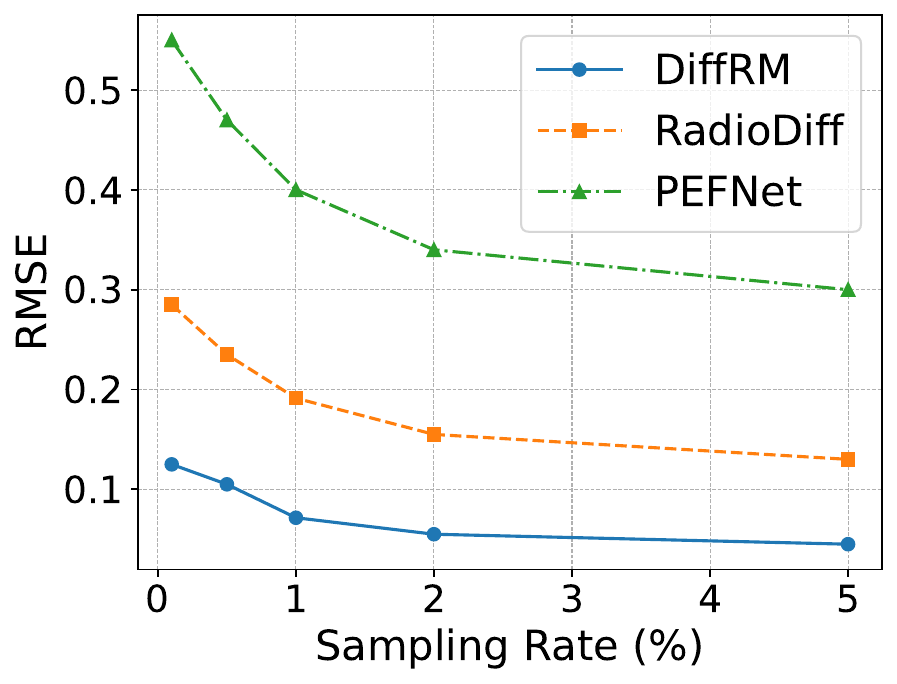}\label{fig: obs_a}}
    \hfil
    \subfloat[SSIM]{\includegraphics[width = 2.0in]{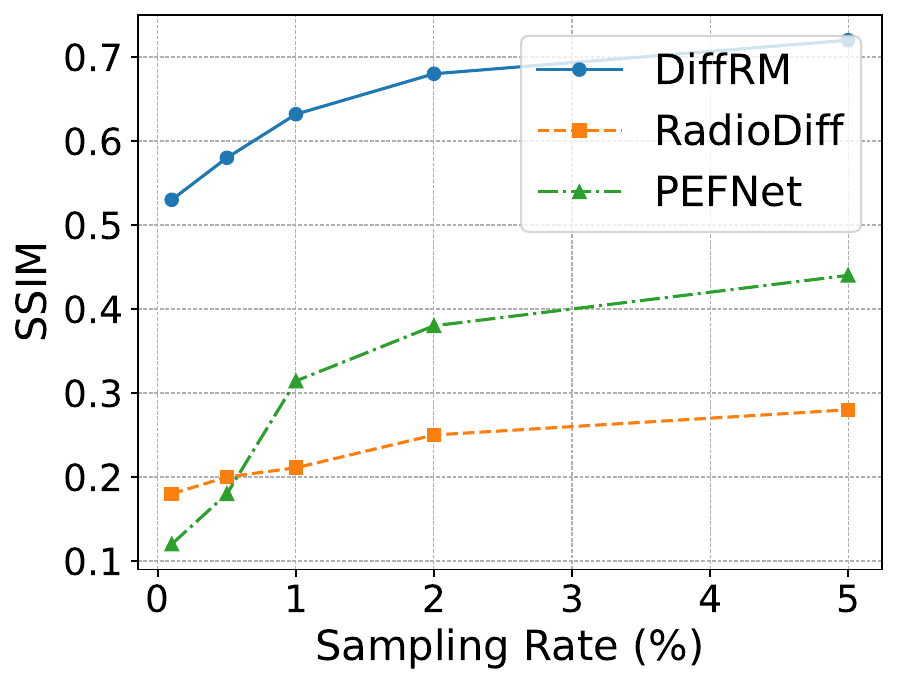}\label{fig: obs_b}}
    \hfil
    \subfloat[PSNR]{\includegraphics[width =  2.0in]{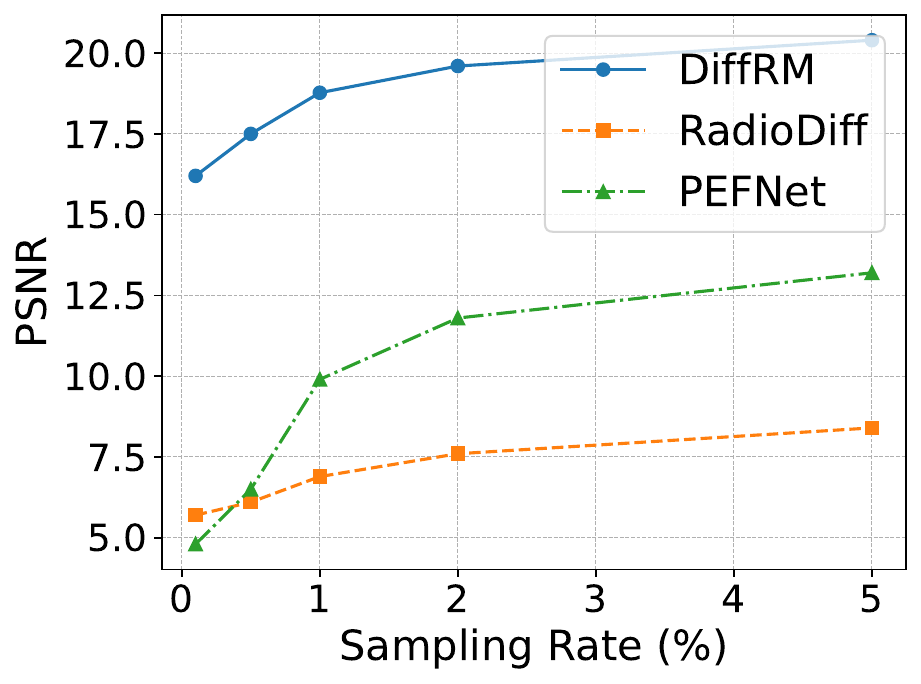}\label{fig: obs_c}}
    \caption{\textcolor{black}{Radio map imputation performance comparison under different observation densities.}}
    \label{fig: obs}
\end{figure*}

\textcolor{black}{
Lastly, we evaluate the robustness of DiffRM under varying observation conditions. Specifically, the model is trained under a fixed sampling rate of 1\% and tested across different sampling densities, including 0.1\%, 0.5\%, 2\%, and 5\%, to assess its generalization capability.
The results are illustrated in Figure~\ref{fig: obs}. It can be observed that DiffRM consistently outperforms baseline methods across all sampling rates in terms of RMSE, SSIM, and PSNR. 
As the sampling rate increases, the performance of all methods improves due to the availability of more observed data. However, DiffRM continues to achieve the best performance, indicating its superior ability to effectively capture complex radio propagation patterns and generalize across different observation densities.
}

\renewcommand{\arraystretch}{1.5}   
\begin{table*}[t]
\footnotesize 
\centering
\caption{Radio map prediction performance in different frequency bands.} 
\vspace{-2mm}
\label{tab:frequency}
\resizebox{1.0\linewidth}{!}
{
    \begin{tabular}{c|c|c|c|c|c|c|c|c|c|c|c|c|c|c|c}
    \hline
    \textbf{Frequency} & \multicolumn{3}{c|}{\textbf{150 MHz}} & \multicolumn{3}{c|}{\textbf{1.5 GHz}} & \multicolumn{3}{c|}{\textbf{1.7 GHz}} & \multicolumn{3}{c|}{\textbf{3.5 GHz}} & \multicolumn{3}{c}{\textbf{22 GHz}} \\
    \hline
    \hline
    \textbf{Method} & {RMSE} & {SSIM} & {PSNR} & {RMSE} & {SSIM} & {PSNR} & {RMSE} & {SSIM} & {PSNR} & {RMSE} & {SSIM} & {PSNR} & {RMSE} & {SSIM} & {PSNR}\\
    \hline
    \textbf{RadioUNet} & {0.1923} & {0.5043} & {11.6932} & {0.1784} & {0.5173} & {11.7388} & {0.1793} & {0.5129} & {12.3021} & {0.1853} & {0.5063} & {11.8273} & {0.1932} & {0.5001} & {11.2537}\\
     \hline
    \textbf{PEFNet} & {0.1735} & {0.4293} & {9.0291} & {0.1673} & {0.4371} & {9.7839} & {0.1702} & {0.4395} & {9.2032} & {0.1784} & {0.4287} & {8.9907} & {0.1880} & {0.4236} & {8.6795}\\
    \hline
    \textbf{RadioDiff} & {0.1634} & {0.2833} & {6.3311} & {0.1517} & {0.2974} & {6.7930} & {0.1528} & {0.2953} & {6.8804} & {0.1672} & {0.2886} & {6.7042} & {0.1698} & {0.2878} & {6.0026}\\
    \hline
    \textbf{DiffRM} & {0.1495} & {0.5909} & {15.9568} & {0.1517} & {0.6148} & {15.9384} & {0.1233} & {0.5970} & {16.9780} & {0.1166} & {0.6079} & {16.6481} & {0.1338} & {0.6108} & {15.9714}\\
    \hline
    \end{tabular}
}%
\end{table*}

\subsection{Ablation Study}

\textcolor{black}{
In this part, we analyze the impact of different module designs on the performance of DiffRM. Specifically, we compare the proposed model with several variants, including: (1) excluding numerical metadata context, (2) removing the textual prompt-based guidance, (3) replacing the CLIP-based text encoder with a learnable embedding, (4) directly feeding the original radio maps into the diffusion model without the VAE-based latent representation, and (5) replacing the sinusoidal encoding of numerical metadata with a learnable embedding.
The results are shown in Table~\ref{tab:ablation}. All variants exhibit performance degradation compared to the original model, demonstrating the effectiveness of each component. Removing either numerical or textual context leads to a noticeable decline, indicating that these modalities provide complementary information for characterizing radio propagation.
For the VAE component, operating directly in the original data space without latent feature extraction yields inferior performance, underscoring the advantage of learning a compact and informative latent representation. In addition, replacing the CLIP-based text encoder with a learnable embedding degrades performance, suggesting that the semantically meaningful representations provided by CLIP further enhance the model’s ability to distinguish between different environmental conditions.
Furthermore, replacing the sinusoidal encoding with a learnable embedding also degrades performance. This indicates that sinusoidal positional encoding is more suitable for modeling frequency- and height-dependent radio propagation characteristics.
}
 
We further investigate the impact of the task-unified masking strategy on DiffRM. Specifically, we design two additional variants, each trained with a single masking strategy, and evaluate them separately on both tasks. 
\textcolor{black}{
The results in Table~\ref{tab:masking} show that task-specific masking strategies achieve competitive performance on their corresponding tasks but generally exhibit degraded performance when applied to the other task.}
In contrast, the proposed unified masking strategy achieves consistently strong performance across both tasks, highlighting its superior generalization capability and necessity.

\begin{table}[t]
\small
\renewcommand{\arraystretch}{1.26}  
\begin{center}
    \caption{Ablation study of radio map prediction performance.} 
    \vspace{-3mm}
    \label{tab:ablation}
    \resizebox{1.0\linewidth}{!}{
    \begin{tabular}{c|c|c|c|c|c}
    \hline
    {\textbf{Method}} & \textbf{MSE} & \textbf{NMSE} & \textbf{RMSE} & \textbf{SSIM} & \textbf{PSNR} \\
    \hline
    \hline
    w.o. metadata & {0.0304} & {0.0691} & {0.1623} & {0.5874} & {16.1047}\\
    \hline
    w.o. prompt & {0.0333} & {0.0762} & {0.1684} & {0.5803} & {15.8929}\\
    \hline
    w. learnable textual emb & {0.0298} & {0.0652} & {0.1402} & {0.5935} & {16.0523} \\
    \hline
    w.o. VAE & {0.0318} & {0.0734} & {0.1670} & {0.5822} & {15.9043}\\
    \hline
    w.o. sinusoidal emb & {0.0297} & {0.0634} & {0.1369} & {0.5942} & {16.1836} \\
    \hline
    DiffRM & {0.0269} & {0.0589} & {0.1279} & {0.6014} & {16.5147} \\
    \hline
    \end{tabular}
    }
    \vspace{-3mm}
\end{center} 
\end{table}

\begin{table}[t]
\small
\renewcommand{\arraystretch}{1.2}
\begin{center}
\caption{\textcolor{black}{Ablation Study on the effect of task-unified masking strategy for radio map construction.}}
\vspace{-3mm}
\label{tab:masking}
\resizebox{\linewidth}{!}{
\begin{tabular}{c|ccc|ccc}
\hline
\multirow{2}{*}{\textbf{Method}} 
& \multicolumn{3}{c|}{\textbf{Prediction}} 
& \multicolumn{3}{c}{\textbf{Imputation}} \\
\cline{2-7}
& {RMSE} & {SSIM} & {PSNR} & {RMSE} & {SSIM} & {PSNR}\\
\hline
\hline
prediction masking & {0.1265} & {0.5930} & {16.6204} & {0.1324} & {0.6903} & {15.3823} \\
\hline
imputation masking & {0.1713} & {0.5693} & {14.0392} & {0.0708} & {0.6312} & {18.8103} \\
\hline
DiffRM & {0.1279} & {0.6014} & {16.5147} & {0.0715} & {0.6319} & {18.7736}  \\
\hline
\end{tabular}
}
\vspace{-3mm}
\end{center}
\end{table}

\section{Related Work} 
\label{sec5}
In this section, we provide an overview of existing studies on radio map construction, which can be categorized into two primary types: model-based and model-free approaches.

Model-based methods usually leverage specific signal propagation models and express the received signal PSD as a combination of active transmitters, which are either formulated by explicit functions based on prior knowledge or represented in terms of sensor parameters. 
For instance, Lee~\emph{et al.}~\cite{lee2012voronoi} introduced a log-distance path loss model (LDPL) tailored for single narrowband WiFi transmitters, where the signal attenuation is modeled as a logarithmic function of distance. 
Bazerque~\emph{et al.}~\cite{bazerque2011group} proposed an interpolation method using thin-plate spline kernels to estimate the power gain, enabling flexible surface fitting of the power field from sparse measurements. 
Other representative model-based approaches include a parallel factor analysis (PARAFAC)-based framework proposed by Fu~\emph{et al.}~\cite{fu2015factor}, which decomposes the radio map tensor into latent factors to capture structural dependencies across multiple dimensions, and the Fixed Rank Kringing (FRK) algorithm introduced by Braham~\emph{et al.}~\cite{braham2016fixed}, which can give accurate coverage predictions on the whole area of interest while reducing the computational complexity.
Since model-based methods rely on the functional approximation properties of the model to fit the current propagation behavior, they typically exhibit superior performance in smooth, regular environments. However, the inappropriate usage of radio propagation models may limit the accuracy of these methods. Especially in complicated and irregular scenarios, the received power distribution is more sensitive to environmental information, which the signal propagation model is unable to capture extensively.

In contrast, model-free methods typically explore more surrounding information rather than assume a specific propagation model. In general, they can be further separated into interpolation-based and learning-based approaches. Interpolation-based models express the PSD at a particular location as a combination of neighborhood measurements. A typical example is inverse distance weighted (IDW) interpolation~\cite{kuo2010discriminant}, which assumes the power gain is inversely proportional to the distance between transmitters and receivers. Radio basis function (RBF) interpolation~\cite{krumm2003minimizing} is a nonlinear alternative, where different kernels, including Gaussian, multiquadrics, and spline, are applied in the method. Another classic approach is the Kringing algorithm~\cite{boccolini2012wireless}, where the PSD is updated by minimizing the variance of the estimation error. Gaussian process regression (GPR) further enhances the modeling capacity by employing compound kernels and the K-Nearest Neighbor localization algorithm for predicting received signal strength~\cite {sun2018augmentation}. Different from explicit interpolation, learning-based models aim to learn a direct mapping from the input geometric features of the environment to the corresponding radio map outputs. This input-output relationship is typically modeled using a neural network (NN), which effectively captures the inherent radio propagation patterns and dependencies. For example, some researchers employ the UNet structure~\cite{ronneberger2015u} as the backbone of their models for radio map prediction~\cite{levie2021radiounet,lee2024scalable,jaensch2024radio}, where the building map and base station location are regarded as inputs. While Jiang~\emph{et al.}~\cite{jiang2024physics} proposed PEFNet, which merges physical knowledge with the UNet neural network to estimate path loss. Additionally, more complex architectures, such as graph neural networks (GNNs), with stronger feature extraction capabilities, have been applied in radio map construction~\cite{chen2023graph,zhang2023rme}. Moreover, the diffusion-based generative model is also employed in the RadioDiff model~\cite{wang2024radiodiff}, where the transmitter position and environmental features are used as prompts for conditional generation.

\section{Conclusion} 
\label{sec6}
In this paper, we propose a generative foundation model, DiffRM, for constructing radio maps across diverse scenarios, including multiple terrain types, frequency bands, and sampling heights. 
By leveraging diffusion models as the backbone, our method effectively captures the probabilistic distribution of received signal strength shown in radio maps. 
Additionally, we have reformulated radio map prediction and imputation into a joint task via masking strategies, enabling our model to learn a general inference mechanism.
Furthermore, multimodal contexts are encoded into latent representations via modality-specific encoders to guide the construction process. 
Extensive experiments on a rich radio map dataset have demonstrated the superior generalization ability of our model compared to baseline approaches.

\textcolor{black}{
Despite these promising results, limitations remain. Due to the limitations of the currently available datasets, weather conditions are categorized into only a few coarse-grained classes, and detailed weather information is unavailable. 
As weather conditions influence radio propagation characteristics, incorporating finer-grained weather information could further improve modeling accuracy. Future work will focus on collecting more detailed weather data to enable more precise weather-aware radio map construction.
}


\bibliographystyle{IEEEtran}
\bibliography{ref}

\balance

\end{document}